\pdfoutput=1
\documentclass{AISB2008}
\usepackage{times}
\usepackage{graphicx}
\usepackage{latexsym}
\usepackage{url}
\usepackage{amsmath}
\usepackage{amssymb}
\usepackage{caption}
\usepackage{tabularx}
\usepackage{floatrow}

\begin{document}
	
	\title{A Genetic Programming Framework for 2D Platform AI}
	
	\author{ Swen E. Gaudl\institute{MetaMakers Institute,
			UK, email: swen.gaudl@gmail.com} }
	
	\maketitle
	\bibliographystyle{AISB2008}
	
	\begin{abstract}
		There currently exists a wide range of techniques to model and evolve artificial players for games. Existing techniques range from black box neural networks to entirely hand-designed solutions. In this paper, we demonstrate the feasibility of a genetic programming framework using human controller input to derive meaningful artificial players which can, later on, be optimised by hand. The current state of the art in game character design relies heavily on human designers to manually create and edit scripts and rules for game characters. To address this manual editing bottleneck, current computational intelligence techniques approach the issue with fully autonomous character generators, replacing most of the design process using black box
		solutions such as neural networks or the like. Our GP approach to this problem creates character controllers which can be further authored and developed by a designer it also offers designers to included their play style without the need to use a programming language.
		This keeps the designer in the loop while reducing repetitive manual labour. Our system also provides insights into how players express themselves in games and into deriving appropriate models for representing those insights. We present our framework, supporting findings and open challenges.
	\end{abstract}
	
	\section{Introduction}
	The design of intelligent systems is a complex task which in itself can benefit from the application of AI techniques. Here we present a system that offers the developer the option to mine human behaviour and include it into the system to create better Game AI. We detail a genetic programming (GP) system that generalises from and improve upon human game play.  More importantly, the resulting representations are amenable to further authoring and development.
	We discuss our GP system for evolving game characters by utilising recorded human play. The system uses the platformerAI toolkit, detailed in section \ref{sec:env}, and the \textsc{Java} genetic algorithm and genetic programming package (JGAP) \cite{JGAP2005}. \textsc{JGAP} provides a system to evolve computer programs and their representations as decision tree when given a set of command genes, a fitness function, a genetic selector and an interface to the target application. Once the system it set up by including those components, it generates artificial players by creating and evolving \textsc{Java} program code which is fed into the \textsc{platformerAI} toolkit and evaluated using our fitness function which is detailed in \cite{Gaudl2015}.
	
	The rest of this paper is organised as follows. In section \ref{sec:bg} we describe how our system derives from and improves upon the start of the art. Section \ref{sec:eval} describes our system and its core components, including details on our the design of fitness functions. We conclude our work by describing our findings and possible open challenges.
	
	\section{Background \& Related Work} \label{sec:bg}
	In practice, making a good game is achieved by a good concept and long iterative cycles in refining mechanics and visuals, a process which is resource consuming. It requires a large number of human testers to evaluate the qualities of a game. Thus, analysing tester feedback and incrementally adapting games to achieve better play experience is tedious and time-consuming. Reducing some part of the laborious work is where our approach comes into play by trying to minimise development, manual adaptation and testing time, yet allow the developer to remain in full control.
	
	\emph{Agent Design} was initially no more than creating 2D shapes on the screen, e.g. the aliens in \textsc{SpaceInvaders}. Due to early hardware limitations, more complex approaches were not feasible. With more powerful computers it became feasible to integrate more complex approaches such as finite state machines (FSMs). 
	In 2002 Isla introduced the \textsc{BehaviourTree (BT)} for the game Halo, later elaborated by Champandard \cite{Champandard2003}. BT uses a directed acyclic graph to represent the reasoning process within the game logic. It integrates hierarchical structures as well offering the system to scale based on the requirements but does not have the same disadvantages of FSMs, namely the exponential amount if transition checks required to verify the functionality of the FSM. BT has become the dominant approach in the industry. BTs can be represented as a combination of a decision tree (DT) using a pre-defined set of node types.  
	A related academic predecessor of the BT were the \textsc{Posh} dynamic plans of \textsc{Bod} \cite{BrysonIJCAI01, Gaudl2013}. 
	
	\emph{Generative Approaches}  build models to create better and more appealing agents. To achieve their goal, a generative agent uses machine learning techniques to increase its capabilities by testing and updating its components. Using data derived from human interaction with a game---referred to as human play traces---can allow the game to act on or \textit{re-act} to input created by the player. By training on such data, it is possible to derive models able to mimic certain characteristics of players \cite{Holmgard2014,Ortega201393} . 
	One obvious disadvantage of this approach is that the generated model only learns from the behaviour exhibited in the data provided to it. Thus, interesting behaviours are not accessible because they were never exhibited by a player.
	
	In contrast to other generative agent approaches \cite{Perez2011, Togelius2012, Ortega201393} our system combines features which allow the generation and development of truly novel agents. Thus, the system presents the first use of un-authored recorded player input as direct input into our fitness function. It allows the specification of agents only by playing. The second feature of the system is that our agents are actual programs in the form of either \textsc{Java} code or decision tree representations which can be altered and modified after evolving into a desired state, creating a white box solution. While \cite{Stanley02} use neural networks (NN) to create better agents and enhance games using Neuroevolution, we utilise genetic programming \cite{poli2008field} for the creation and evolution of artificial players in human readable and modifiable form. The most comparable approach is that of \cite{Perez2011} which use grammar based evolution to derive BTs given an initial set and structure of sub\-trees. In contrast, we start with a clean slate to evolve our agents as directly executable programs.
	
	\section{Setting and Environment} \label{sec:env} 
	Evolutionary algorithms have the potential to solve problems in
	vast search spaces, especially if the problems require multi-parameter optimisation \cite[p.2]{schwefel1993evolution}. For those problems, humans are generally outperformed by programs \cite{smit2009comparing}. 
	Our GP approach uses a pool of program chromosomes $P$ and evolves those in the form of decision trees (DTs) exploring the possible solution space. For our experiments the \textsc{platformerAI} toolkit (\url{http://www.platformersai.com}) was used which is entirely written in \emph{Java} and freely available. It consists of a 2D platformer game, similar to existing commercial products and contains modules for recording a player, controlling agents and modifying the environment and rules of the game.
	
	The \textit{Problem Space} is defined by all actions an agent can perform.
	Within the game, agent $A$ has to solve the complex task of selecting the appropriate action each given frame. The game consists of $A$ traversing a level which is not fully observable. A level is 256 spatial units long, and $A$ should traverse it left to right. Each level contains objects which act in a deterministic way. Some of those objects can alter the player's score, e.g. coins. Those bonus objects present a secondary objective. The goal of the game, move from start to finish, is augmented with the objective of gaining points. $A$ can get points by collecting objects or jumping onto enemies.
	To make it comparable to the experience of similar commercial products
	we use a realistic time frame in which a human would need to solve a
	level, 200 time units. The level observability is limited to a $6\times6$ grid centred around the player, cf. \cite{Perez2011}. The restriction to a smaller grid is only necessary to reduce the number of generations the system needs to converge towards good results as the grid size has an exponential affect on the convergence time. 
	
		\begin{figure}[ht]
		\includegraphics[width=.6\textwidth]{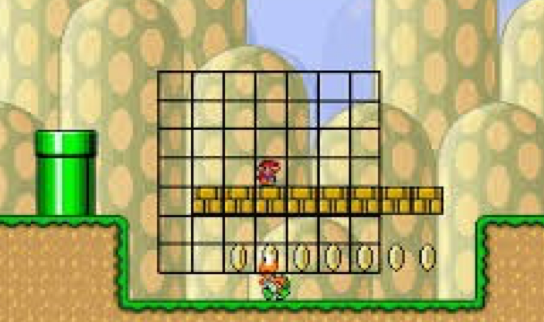}
		%{\caption{The visual representation of a part of the DT for a generated agent, evolved using \textsc{PBF} after 694 generations and able to win the first level. JGAP uses this DT in the form of java code to control an agent.} \label{fig:evolved_pbf_plan}}
		%\vspace*{-6ex}
		\caption{A visual representation of the \textsc{platformersAI} toolkit with the vision grid around the agent.} 
		\label{fig:agent-framework}
	\end{figure}

	\textit{Agent Control} within the platformersAI toolkit is handled through a 6-bit vector $C$: $left$, $right$, $up$, $down$, $jump$ and $shoot | run$. The vector is required each frame, simulating an input device to control the agent in Figure\ref{fig:agent-framework}. However, some actions span more than one frame. This is a simple task for a human but quite complex to learn for an agent. One such example, the high jump, requires the player to press the jump button for multiple frames. Those long action sequences mean that the agent needs to anticipate future events and actions to trigger actions spanning multiple reasoning cycles.
	Our system has genes for each of the elements of $C$ plus 14 additional genes formed of five gene types: sensory information about
	the level or agent, executable actions, logical operators, numbers and
	structural genes. All those are combined at execution time into a
	chromosome represented as a DT using the grammar underlying the
	\textsc{Java} language. Structural genes allow the execution of $n$
	genes in a fixed sequence, reducing the combinatorial freedom provided by \textsc{Java}. Our system uses the JGAP framework, which allows us to add new genes to enrich the search space and the agent capabilities by writing self-contained \textsc{Java} methods and adding them to the Agent class. However, adding more genes increases the search space resulting potentially in longer conversion times. 
	
	\begin{table}
		\begin{tabularx}{\linewidth}{ |X|X|}
			\hline
			Parameter & Value \\
			\hline
			Initial Population Size  & 100  \\
			\hline
			Selection  & Weighted Roulette Wheel  \\
			\hline
			Genetic Operators  & Branch Typing CrossOver and Single Point Mutation  \\
			\hline
			Initial Operator probabilities  & 0.6 crossover, 0.2 new chromosomes, 0.01 mutation, fixed  \\
			\hline
			Survival  & Elitism  \\
			\hline
			Function  Set & $ifelse$, $not$, $\&\&$, $||$, $sub$, $IsCoinAt$, $IsEnemyAt$, $IsBreakAbleAt$, $\dots$  \\
			\hline
			Terminal Set  & Integers [-6,6], $\leftarrow$, $\rightarrow$, $\downarrow$, $IsTall$, $Jump$, $Shoot$, $Run$  $Wait$, $CanJump$, $CanShoot$, $\dots$ \\
			\hline
		\end{tabularx}
		%\vspace*{-6ex}
		\caption{GP parameters used in our system.}
		\label{tbl:para}
	\end{table}
	
	\section{Fitness Evaluation}
	The evaluation is done \label{sec:eval} in our system using the Gamalyzer-based play trace metric which determines the
	fitness of individual chromosomes based on human traces as an
	evaluation criterion, see \cite{Gaudl2015}. 
	For finding optimal solutions to a problem, statistical fitness functions offer near-optimal results when optimality can be defined. A near-best solution for the problem space of finding the optimal way through a level in the platformersAI toolkit was given by Baumgarten \cite{Togelius2010} using the $A^{*}$  algorithm. This approach produces agents who are extremely good at winning the level within a minimum amount of time but at the same time are clearly distinguishable from actual human players.  Contrasting the goal of finding optimal solutions, we are interested in understanding and modelling human-like or human-believable behaviour in games. Thus, using statistical functions is difficult, as there currently is no known algorithm for measuring how human-like behaviour is; identifying this may even be computationally intractable. For games and game designers a less distinguishable approach is normally more appealing---based on our initial assumptions. Additionally, having an approach which produces readable and amenable representations of the behaviour might not just aid its understanding but might offer different insights into the design of the game as well.
	
	Based on the biological concept of selection, all evolutionary systems
	require some form of judgement about the quality of a specific
	individual---the fitness value of the entity. 
	Within our framework, agents are evaluated after each run of an entire level of the game as intermittent evaluation of games where actions can span multiple cycles is difficult to evaluate. Within the original \textsc{JGAP} framework evaluation can be done at arbitrary times but it an important consideration that the evaluation (running the program to receive a result) is normally the most expensive cost within a GP.
	
	In table \ref{tbl:para} the settings we use for GP within our framework are given. As a selection mechanism, the weighted roulette wheel is used which attributes each chromosome a position and then weights all chromosomes according to their fitness giving fitter individuals slightly more space. We additionally preserve the fittest individual of a generation.  Preserving the best individual is crucial as mutation can be destructive to the chromosome We use single point tree branch crossover on two selected parent chromosomes and expose the resulting child to a single point mutation before it is put into the new generation. We also add 20\% new randomly generated chromosomes to the pool to bring in some "fresh blood" or to be more precise keep the pool from stopping in a homogeneous state. Even through mutation is potentially destructive, it helps exploring the vast gene space better than relying entirely on the cross-over operation. However, within our experiments \cite{Gaudl2015}, using the more stable cross-over as the main driving force for the evolution gave better and more reliant results than switching entirely to random exploration using a stronger mutation coefficient. 
	
	\begin{figure}[ht]
		\includegraphics[width=\linewidth]{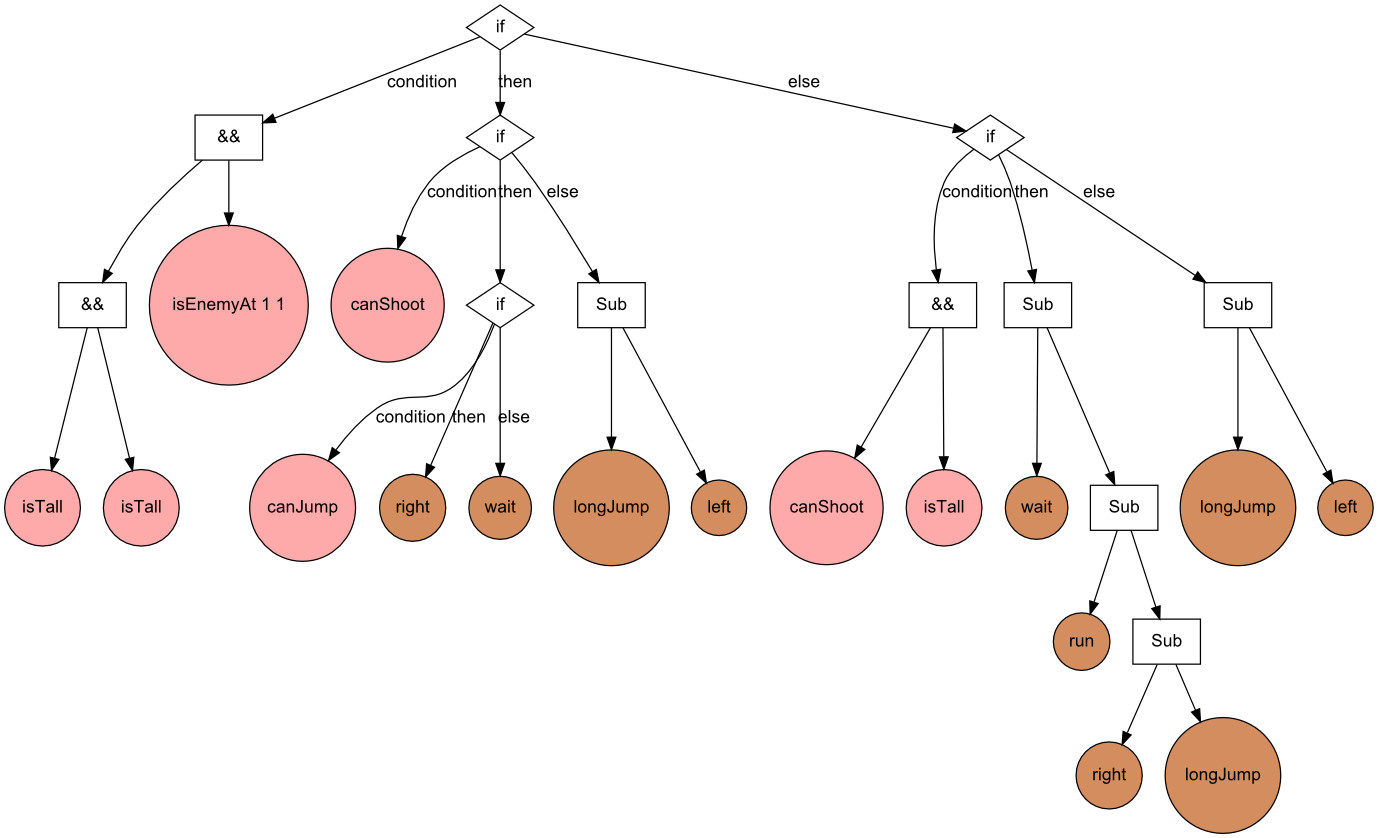}
		%\vspace*{-8ex}
		\caption{An evolved agent after 694 generations, represented as decision tree by our system.} 
		\label{fig:evolved_agent}
	\end{figure}
	
	In Figure \ref{fig:evolved_agent} one of the resulting agents is presented in its DT form. The visual representation was generated by the system using \textsc{Graphviz} (\url{http://www.graphviz.org/}). As the aim of our approach was to derive meaningful representations of agent behaviour, visual representation of the result is of utmost importance. Using the rendered DT allows a designer to either alter the agent or to understand why it behaved in a certain way. 
	
	\section{Findings \& Open Challenges}
	\label{sec:results}
	Using our experimental configuration and the \textsc{PBF} fitness function \cite{Gaudl2015} we are now able to execute, evaluate and compare platformerAI agents against human traces. 
	Using human play traces to drive the evolution resulted in agents which are able to beat some but not all of the test levels. However, there is still potential using different ways to integration human knowledge into the evaluation. 
	The \textsc{JGAP} framework proved to a useful and easy to use and robust framework for developing genetic programs, even though it has some weaknesses compared to other frameworks. If you care for running the GP on a cluster you might decide to use a different framework which offers better support for spitting up both the evaluation of chromosomes and the handling large data structures. 
	Most of the GP systems let you also run or communicate external libraries. In our case, we included the \textsc{platformersAI} toolkit to evaluate our agents. This toolkit does not support parallel instantiations of multiple levels well but can be tweaked easily and offers also support for using a genetic approach to evolve levels.
	A next step would be to investigate the generated modifiable programs further and analyse their benefit in understanding players better. However, our current solution already offers a way to design agents for a game by simply playing it and creating learning agents from those traces.
	Other possible directions could be the comparison of different fitness functions and how different interpretations of human play input might affect the convergence rate of agents within our framework. Our current agent model consists of an un\-weighted tree representation containing program genes.  Currently subtrees are not taken into consideration when calculating the fitness of an individual. By including those weights it would be possible to narrow down the search space of good solutions for game characters dramatically, also potentially reducing the bloat of the DT. So, to enhance the quality of our reproduction component we believe it might be interesting to investigate the applicability of behavior-programming for GP (BPGP) \cite{krawiec2014behavioral} into our system.

	\bibliography{biblio/sweneg,biblio/swen-ai-lit,biblio/swen-game-lit,biblio/swen-phil-lit,biblio/jjb}  % sigproc.bib is the name of the Bibliography in this case
	% You must have a proper ".bib" file
	%  and remember to run:
	% latex bibtex latex latex
	% to resolve all references
	%
	% ACM needs 'a single self-contained file'!
	%
	%APPENDICES are optional
	%\balancecolumns
\end{document}